\documentclass[journal,onecolumn]{IEEEtran}

\usepackage{epsfig}
\usepackage{graphicx}
\usepackage{amsmath}
\usepackage{amssymb}
\usepackage{commath}
\usepackage{bbding}
\usepackage{cite}
\usepackage{cleveref}
\usepackage{booktabs,multirow}
\usepackage{times}
\usepackage{epstopdf}

\usepackage{mathtools}

\begin{document}

%--------------------------------------------------------------------------------------------- 
%--------------------------------------------------------------------------------------------- 
%--------------------------------------------------------------------------------------------- 
\title{Learning Semantics for Image Annotation}
%
%
%--------------------------------------------------------------------------------------------- 
%--------------------------------------------------------------------------------------------- 
\author{Hassan Foroosh}
\author{Amara Tariq\thanks{Amara Tariq was with the Department of Computer Science, University of Central Florida, Orlando, FL, 32816 USA at the time this project was conducted (email: amara\_tariq@knights.ucf.edu).} and Hassan Foroosh\thanks{Hassan Foroosh is with the Department of Computer Science, University of Central Florida, Orlando, FL, 32816 USA (email: foroosh@cs.ucf.edu).}
}

\maketitle

%--------------------------------------------------------------------------------------------- 
%--------------------------------------------------------------------------------------------- 
\begin{abstract}
Image search and retrieval engines rely heavily on textual annotation in order to match word queries to a set of candidate images. A system that can automatically annotate images with meaningful text can be highly beneficial for such engines. Currently, the approaches to develop such systems try to establish relationships between keywords and visual features of images. In this paper, We make three main contributions to this area: (i) We transform this problem from the low-level keyword space to the high-level semantics space that we refer to as the ``{\em image theme}'', (ii) Instead of treating each possible keyword independently, we use latent Dirichlet allocation to learn image themes from the associated texts in a training phase. Images are then annotated with image themes rather than keywords, using a modified continuous relevance model, which takes into account the spatial coherence and the visual continuity among images of common theme. (iii) To achieve more coherent annotations among images of common theme, we have integrated ConceptNet in learning the semantics of images, and hence augment image descriptions beyond annotations provided by humans. Images are thus further annotated by a few most significant words of the prominent image theme. Our extensive experiments show that a coherent theme-based image annotation using high-level semantics results in improved precision and recall as compared with equivalent classical keyword annotation systems.
\end{abstract}


%--------------------------------------------------------------------------------------------- 
%--------------------------------------------------------------------------------------------- 
\section{Introduction}

With the advancement in information search and retrieval techniques, annotation of images with keywords has been a popular area of research \cite{Tariq_etal_2017,Tariq_etal_2017_2,tariq2013exploiting,tariq2015feature,tariq2014scene}. Annotations often contain content-related information such as objects and shapes/patterns present in the scene \cite{Cakmakci_etal_2008,Cakmakci_etal_2008_2,Zhang_etal_2015,Lotfian_Foroosh_2017,Morley_Foroosh2017,Ali-Foroosh2016,Ali-Foroosh2015,Einsele_Foroosh_2015,ali2016character,Cakmakci_etal2008,damkjer2014mesh,Shu_etal_2016,Milikan_etal_2017,Millikan_etal2015,shekarforoush2000multi,millikan2015initialized}, structural scene models \cite{Junejo_etal_2013,bhutta2011selective,junejo1dynamic,ashraf2007near,Junejo_etal_2007,Junejo_Foroosh_2008,Sun_etal_2012,junejo2007trajectory,sun2011motion,Ashraf_etal2012,sun2014feature,Junejo_Foroosh2007-1,Junejo_Foroosh2007-2,Junejo_Foroosh2007-3,Junejo_Foroosh2006-1,Junejo_Foroosh2006-2,ashraf2012motion,ashraf2015motion,sun2014should}, humans and their actions \cite{Shen_Foroosh_2009,Ashraf_etal_2014,Ashraf_etal_2013,Sun_etal_2015,shen2008view,sun2011action,ashraf2014view,shen2008action,shen2008view-2,ashraf2013view,ashraf2010view,boyraz122014action,Shen_Foroosh_FR2008,Shen_Foroosh_pose2008,ashraf2012human},  or information about the location of the image \cite{Junejo_etal_2010,Junejo_Foroosh_2010,Junejo_Foroosh_solar2008,Junejo_Foroosh_GPS2008,junejo2006calibrating,junejo2008gps}. As a result there is often a semantic gap between the annotations and the image, since the annotations are dealing with content and seldom with context/theme. From an application point of view, search engines and retrieval systems rely on annotation with textual data to match images with textual queries. Such queries may be used in applications such as image editing and post-production \cite{Cao_etal_2005,Cao_etal_2009,shen2006video,balci2006real,xiao20063d,moore2008learning,alnasser2006image,Alnasser_Foroosh_rend2006,fu2004expression,balci2006image,xiao2006new,cao2006synthesizing}, matching places by alignment \cite{Foroosh_etal_2002,Foroosh_2005,Balci_Foroosh_2006,Balci_Foroosh_2006_2,Alnasser_Foroosh_2008,Atalay_Foroosh_2017,Atalay_Foroosh_2017-2,shekarforoush1996subpixel,foroosh2004sub,shekarforoush1995subpixel,balci2005inferring,balci2005estimating,foroosh2003motion,Balci_Foroosh_phase2005,Foroosh_Balci_2004,foroosh2001closed,shekarforoush2000multifractal,balci2006subpixel,balci2006alignment,foroosh2004adaptive,foroosh2003adaptive}. 
Images that are of low resolution and low quality may be improved by preprocessing methods \cite{Foroosh_etal_2002,Foroosh_2005,Balci_Foroosh_2006,Balci_Foroosh_2006_2,Alnasser_Foroosh_2008,Atalay_Foroosh_2017,Atalay_Foroosh_2017-2,shekarforoush1996subpixel,foroosh2004sub,shekarforoush1995subpixel,balci2005inferring,balci2005estimating,foroosh2003motion,Balci_Foroosh_phase2005,Foroosh_Balci_2004,foroosh2001closed,shekarforoush2000multifractal,balci2006subpixel,balci2006alignment,foroosh2004adaptive,foroosh2003adaptive}, in order to identify content, or may benefit from camera pose and motion quantification methods for scene modeling \cite{Cao_Foroosh_2007,Cao_Foroosh_2006,Cao_etal_2006,Junejo_etal_2011,cao2004camera,cao2004simple,caometrology,junejo2006dissecting,junejo2007robust,cao2006self,foroosh2005self,junejo2006robust,Junejo_Foroosh_calib2008,Junejo_Foroosh_PTZ2008,Junejo_Foroosh_SolCalib2008,Ashraf_Foroosh_2008,Junejo_Foroosh_Givens2008,Lu_Foroosh2006,Balci_Foroosh_metro2005,Cao_Foroosh_calib2004,Cao_Foroosh_calib2004,cao2006camera}. However, such preprocessing and restoration methods cannot help in extracting high-level semantics about the images, i.e. image themes. On the other hand, most images uploaded on the web have some sort of accompanying textual information e.g. image caption, neighboring text on the same web-page, etc. These forms of textual information can be extremely noisy, or may be dependent on user input like in image caption. The main task of automatic image annotation (AIA) is thus to develop a system to automatically generate keywords for input images. Generated keywords need to be meaningful enough to be used to match images to queries. Most of the previously developed approaches for AIA have been tested over the Corel5K dataset, used initially by Duygulu et al. \cite{corel5k}.  Most popular techniques for AIA have used translation models from natural language processing (NLP) to establish relationships between low-level visual features and keywords.\\

Many of the previously popular techniques have the shortcoming of treating each keyword independently of all the other keywords. It is evident that keywords used as annotation for an image are heavily correlated to each other. For example, if a certain image has been annotated with `people' , `sand'  and `water', chances of `beach' being another correct annotation are much higher than that of `snow'. Any system which completely ignores this quality of correlation between keywords actually misses an important piece of evidence. Over the years, several papers have been published, attempting at incorporating correlation between keywords in automatic image annotation \cite{AIAlangmodel,wordcorrb}. The basis of techniques used to exploit correlation between keywords range from expectation maximization to incorporation of natural language processing (NLP) tools such as WordNet \cite{AIAlangmodel,wordcorrWN}. In this paper, we attempt to exploit the correlation between keywords by using higher level semantics of available annotations. Our technique is based on image theme modeling using Latent Dirichlet Allocation. Thus, we have transformed the problem of low-level keyword annotation to high-level {\em image theme annotation}.\\

Our motivation is based on the fact that low-level visual features may not provide sufficient visual cues for each object in the image to be identified separately, and hence used for annotation. Objects can by partially obstructed from the view or may occupy a too-small size in the image to generate enough evidence in the form of low-level visual features \cite{AIAlangmodel,wordcorrb}. Overall, the {\em semantic gap} between visual features and meaningful annotations remains unbridgeable. But all these visual features combined with their spatial information can provide enough information to identify a theme for an image. The question is therefore how we can find a good annotation for that theme. Since the Corel5K dataset uses limited keyword annotations for each image, we decided to use the IAPR TC 12 dataset, because it provides each image with more complete descriptions in sentences. We have used Latent Dirichlet Allocation (LDA) to perform image theme modeling over descriptions of these images. LDA provides a means of modeling all image themes present in the textual description of images in the form of word distribution. The image theme models generated through this process are based on words used in documents provided for training of the system. Therefore, these image themes implicitly employ correlation between words. We use these image themes for annotation of our training dataset, and then use relevance model to establish the relationship between the annotated images and the visual themes. We thus transform the problem of relevance between keywords and visual objects to image themes and visual objects. Later, we tag each image with the most significant words of each image theme model associated with the image. Thus our final output is similar to other popular image annotation systems and can be directly compared against them.\\

An important advantage of annotation with image themes versus keywords is that image themes can be elaborated using NLP resources like the ConceptNet \cite{CN}. The use of ConceptNet shows that image semantics, when represented by significant words of each image theme, are readily understandable and meaningful for humans. Therefore, annotating images with even the top few words of an image theme will be helpful for humans when searching for images.  The elaborated image themes may contain words which are not present in ground-truth image descriptions available with the datasets, but provide additional contextual knowledge about the images. We have demonstrated this effect on a smaller part of the test datasets, through manual tagging.\\
We have used a relevance model, similar to the Continuous Relevance Model (CRM) to annotate images with image themes. Our relevance model has two main distinctive features: (i) we take into account the spatial position of visual features; (ii) We incorporate image clustering based on visual features. Each image cluster contains images with a certain level of similarity between their visual features. The size of the pool of possible image themes for each of these clusters is smaller than the overall number of image themes for the whole dataset. Moreover, these image themes will have some level of similarity between them as they are visually similar images. We show that incorporating spatial information and using clusters provide better performance for image annotation.
%-------------------------------------------------------------------------

\section{Related Work}
The idea of annotating images with keywords has been vastly studied using different approaches in the literature, with most papers using the Corel 5K dataset as their benchmark \cite{corel5k,CMRM,CRM}. All different approaches essentially try to learn the relationships between words and image features. Zhang et al. have provided a comprehensive review of all the popular techniques used for automatic image annotation \cite{review}. \\

Relevance models from machine translation were introduced to solve this problem by Jeon et al. \cite{CMRM}. To apply the relevance models, it is necessary to represent images in terms of visual features in a manner similar to the way documents are represented in terms of word-counts. Therefore, Jeon et al.~ and many other researchers used the bag-of-words approach for image representation, which clusters image features to produce a finite number of visual-words. Blobworld by Carson et al.~ was popularly used for dividing images into meaningful patches of similar color and texture \cite{bbworld}. Lavrenko et al.~ introduced the relevance model in the continuous space named as the continuous-relevance-model (CRM) \cite{CRM}, and showed considerable improvement by removing the constraint of finite number of visual-words. Feng et al. introduced the multiple-bernoulli-relevance model and observed that dividing images into a fixed size grid works better than the complex system of Blobworld \cite{MBRM}.  \\

The annotation problem has been also sometimes treated as a classification problem with class-labels as keywords to be used for annotating images \cite{AIAclass1}. This approach works well with primitive datasets of very small number of keywords. Some attempts have also been made to incorporate language models and natural language processing tools such as WordNet in the process of image annotation \cite{AIAlangmodel,AIAWN}. Some researchers have tried to exploit the correlation between keywords during the process of image annotation, rather than treating each keyword independently of all others \cite{AIAlangmodel,wordcorrb,wordcorrWN}. Latent Dirichlet Allocation based image theme modeling was introduced to produce annotation for news images \cite{TMnews}. In this case, each image is accompanied by a news article, which provides additional information regarding that image. Feng et al.~ worked to establish a similar approach to unify visual and linguistic characteristics of images \cite{UnifiedTM}. Makadia et al.~ conducted a detailed survey of automatic image annotation techniques and arrived at the conclusion that greedy label transfer based approaches can beat complex relevance based algorithms in many cases. They presented two such label transfer based techniques \cite{baseline}.\\

In this paper, we propose a solution to a related but new problem of {\em theme-based image annotation}, where the goal is to annotate images with textual information that model image semantics at a higher level than keywords associated with individual objects in the scene. We generate these image theme models using Latent Dirichlet Allocation (LDA) and each image theme is modeled in terms of word distributions. The process of image theme modeling implicitly employs the correlation between words. Therefore, our system overcomes the shortcoming of treating each keyword independently, while they are actually heavily correlated in human perception. These image themes can be represented by a group of few significant words i.e. words with highest probability values in the corresponding word distribution. These words may also be used to generate proper phrases generated by natural language generation techniques. Our motivation is that these image theme models provide better annotation for images, since they provide contextual information rather than reflecting only the content. Our approach is integrating ConceptNet \cite{CN} in learning the semantics of images to prove commonsense basis of our annotation and ground-truth augmentation beyond the description provided by users. We have used the IAPR TC 12  \footnote{http://imageclef.org/photodata} data set for evaluation. This data set is considerably more challenging than the Corel 5K dataset. \\

In the remainder of this paper, we first discuss the problem of theme-based image annotation and our solution based on a modified CRM and ConceptNet for learning high-level image semantics. We then describe the data sets used to assess our solution, followed by the results from a comprehensive set of experiments to: (i) evaluate the performance of our method, (ii) to compare our results with state-of-the-art annotation methods that use low-level keywords in terms of precision and recall, and (iii) to demonstrate visually how augmenting low-level keywords with high-level image theme concepts can enrich the image semantics captured by annotations. We finish the paper with some concluding remarks.

\section{Image Theme Annotation}

In this section, we discuss the various parts of our overall solution for learning high-level semantics for theme-based annotation of images.
\subsection{Image Theme Modeling}
\label{sn:TM}
Image theme modeling through Latent Dirichlet Allocation (LDA) presented by Blei et al. \cite{LDA} has gained tremendous popularity among natural language processing (NLP) researchers. The basic framework of LDA is a generative probabilistic model, which assumes that documents are random mixtures of latent image themes, while each latent image theme can be represented by a distribution over words. Given a set of documents, LDA assumes: (i) A {\em word} is the basic discrete unit of data. (ii) A {\em document} is a sequence of $N$ words. In our case, the description of each image is a document i.e. a sequence of words. (iii) A {\em corpus} is a collection of $M$ documents. In our case, the collection of descriptions of all images constitutes the corpus. \\

According to Blei et al. , the system assumes the following generative process for generating all documents with $\alpha$ and $\beta$ as system parameters.
\begin{itemize}
\item choose $N$ from a Poisson distribution P($\Psi$)
\item choose $\theta$ from a Dirichlet distribution Dir($\alpha$)
\item  for each of the $N$ words $w_n$
\begin{itemize}
\item  choose the image theme $z_n$ from Multinomial($\theta$)
\item  choose the word $w_n$ from a multinomial distribution conditioned on $z_n$ i.e. P($w_n$/$z_n$ , $\beta$)
\end{itemize}
\end{itemize}
Several simplifying assumptions are made, details of which can be found in the paper by Blei et al.~ \cite{LDA}. The final expression for the joint distribution of an image theme mixture $\theta$ and the sets of image themes \textbf{z} and words \textbf{w} is as follows \cite{LDA}:

\begin{equation}
P(\theta,\textbf{z},\textbf{w}/\alpha,\beta) = p(\theta/\alpha)\prod_{n=1}^{N} p(z_n/\theta) p(w_n/z_n,\beta)
\label{eq:lda}
\end{equation}

The process takes as input a corpus of documents and assumes that the above mentioned generative process was at play when these documents were generated. The system estimates the image theme distribution for each document and word distribution, conditioned on image theme distribution. Blei et al.~ have described a variational inference algorithm to estimate the posterior distribution of the hidden variables given a document.

\begin{table*}[hbtp]
\centering
\caption{Sample word distribution conditioned over image theme distribution generated using LDA over image descriptions from the IAPR data set. These word distributions hint towards three distinct visual themes. }
\label{tb:image themes}
\begin{tabular}{|c|c|}\hline
\textbf{image theme1} & `shorts': $0.1275$ , `cyclist': $0.119$, `helmet': $0.114$  , `jersey':  $0.113$, `cycling': $0.0932$\\\hline
\textbf{image theme2} & `table': $0.2$ , `wooden': $0.17$, `walls': $0.06$  , `restaurant':  $0.052$, `glasses': $0.05$\\\hline
\textbf{image theme3} &  `forest': $0.26$ , `bushes': $0.20$, `dense': $0.12$  , `path':  $0.06$, `vegetation': $0.051$\\\hline
\end{tabular}
\end{table*}

We adopted a basic LDA model from the NLP community to apply to image annotation. For this purpose, we used the IAPR dataset, which provides a complete description of each image in three languages. We restricted our research to the English language only. We employed the basic LDA framework using the description of each image as one document. The image theme distribution for each document tells us which image themes are strongly present in each image description in the training set. The word distribution conditioned over the image theme distribution tells us which words are strongly associated with each of the image themes. Thus, we transform the annotation information from the word to the image theme space. \\

Table \ref{tb:image themes} provides some sample word distributions generated for some image themes. These lists have been ranked according to the strength of the probability values and show only the top few words along with their probability values. The word distributions provided in Table \ref{tb:image themes} strongly hint towards distinct visual themes such as `cycle race', `inside of a restaurant' and `dense forest'. These word distributions support our assumption that image themes correspond to visual themes of images. We use low-level visual features with spatial information to find visual themes in images and later find the relevance between those themes and image themes.

\subsection{Integration of ConceptNet}
\label{sn:CN}
ConceptNet is a freely available commonsense knowledge-base \cite{CN}, which has been popularly used for reasoning tasks on documents. This knowledge-base is basically a semantic network, which connects various words and phrases if they are conceptually related, e.g. ``learn'' , ``teacher'' and ``classroom'' are strongly connected to each other in ConceptNet, although these words do not have standard lexical relationships of synonymy, hypernymy or meronymy between them. The data for this knowledge-base was collected as part of the OpenMind Common Sense (OMSC) project, where internet users were asked to fill in templates of information. For example, a certain template `-- is used for --' can be filled as `KNIFE is used for CUTTING'. This simple template will provide information regarding a certain type of relation between words that has been named as `UsedFor' relation. Relationships between words have not only been extracted directly from templates filled in by users, but also through a complex system of inference using the templates as input. The idea is that commonsense knowledge is possessed by every person. Therefore, contributors to this knowledge base do not need to have some specific qualifications as long as the application interface is easy enough to be understood by an average Internet user. \\

ConceptNet is dedicated to contextual reasoning. This semantic network has about $1.6$ million assertions between $300,000$ nodes. A major portion of these assertions are generic conceptual assertions called k-lines \cite{CN}. Other natural language processing resources like WordNet do not have these conceptual assertions between words, but rely on standard lexical relations of synonymy, hypernymy or meronymy. These conceptual assertions enable ConceptNet to perform reasoning over textual input. Overall, there are twenty different types of assertions between nodes. Every assertion is weighted based on the number of times it occurs in the OMSC corpus, and how well it can be inferred indirectly from other assertions. These weights basically measure relatedness between words, and we have used them to reflect the fact that the top few words in each of the image theme models are strongly related to each other in this commonsense based semantic network. \\

We have used image theme modeling to transform keywords into image themes in a manner similar to image theme modeling used in the NLP community to generate image themes for the documents given as input. Therefore, image themes generated through image theme modeling will be better representatives of semantic contents of the input image data. On the other hand, ConceptNet includes generally acceptable commonsense relationships between words. There may be image themes specific to a data set, which may not be represented with high-confidence in ConceptNet. For example, if a data set contains many images of men wearing blue jackets (and described so in image description), one image theme generated through image theme modeling will have ``men'' , ``blue'', and ``jacket'' as the top three words in terms of the probability distribution. ConceptNet, on the other hand, will not have these three words linked to each other with high confidence, because these words do not represent any generally popular concept.  Still, we have shown that significant words of many image themes generated in the given data set are indeed linked with high-confidence in ConceptNet. Therefore, if an image is annotated with the top few words of an image theme, these annotations will convey a strong hint towards a common-sense acceptable concept or theme, providing thus a commonsense basis for our idea of image theme annotation of images. Moreover, the API of conceptnet2.0 provides a tool named `Projection', which takes as input a list of words and returns an extended list of words ranked according to their measure of aggregate relatedness to all words in the input. We provide this tool with the annotation generated by using the significant words of the associated image themes, and then use the output list of words to augment the annotations generated for a specific image.

%-------------------------------------------------------------------------

\subsection{Modified Relevance Model}
\label{sn:RM}
Relevance models are basically statistical formalisms to model the relationship between contents of two corpora. These models have been particularly popular in the natural language processing community for tasks such as machine translation, where it is necessary to establish the relationships between two corpora of text in different languages. In the case of text or language processing, data is usually represented in the form of word counts and is in the discrete domain. Lavrenko et al.~ transformed the relevance model from machine translation to adapt to visual features in the continuous space, and named the new model as the Continuous Relevance Model (CRM) \cite{CRM}. Suppose $\mathcal{T}$ is the set of training images, $J$ is a member of $\mathcal{T}$, $J$ is represented in the form of image regions $r_{J} = \{r_1,r_2...r_n\}$ and the annotation for $J$ are $w_{J} = \{w_1,w_2...w_m\}$. Lavrenko et al.~ assume that (i) the words in $w_{J}$ are i.i.d. random samples from the underlying distribution of $P(./J)$. (ii) the regions in $r_{J}$ correspond to the generator vector $g_1,g_2...g_ n$, generated by some function $G(r)$, with $P_{\mathcal{R}}(r_i/g_i)$, which is independent of $J$. (iii) the generator vectors are also i.i.d. random samples from the multi-variate density function $P_G(./J)$. \\ 

Now if $A$ is an image not in $\mathcal{T}$, with regions $r_{A} = \{r_1,r_2...r_{n_A}\}$ and some arbitrary sequence of words  $w_{B} = \{w_1,w_2...w_{N_B}\}$, the goal is to find the joint distribution of observing $A$ with words in $w_B$. CRM computes an expectation over all images in $\mathcal{T}$ to estimate this joint distribution. The overall process of jointly generating $r_A$ and $w_B$ is as follows \cite{CRM}:
\begin{itemize}
\item Pick a training image $J$ from the set $\mathcal{T}$ with some probability $P_{\mathcal{T}}(J)$
\item For $b=1...n_B$, pick a word $w_b$ from the multinomial distribution $P_{\mathcal{V}}(./J)$. $\mathcal{V}$ denotes the overall vocabulary set.
\item For $a=1...n_A$
\begin{itemize}
\item Sample the generator vector $g_a$ from $P_G(./J)$
\item Pick the region $r_a$ with probability $P_{\mathcal{R}}(r_a / g_a)$
\end{itemize}
\end{itemize}
The formal expression for the joint probability of observation \{$r_A , w_B$\} follows the Drichlet posterior expectation with parameters $\{\mu p_v + N_{v,J}: v \epsilon \mathcal{V}\}$  
\begin{equation}
\begin{split}
P(r_A,w_B) = \sum_{J \epsilon \mathcal{T}} P_{\mathcal{T}}(J) \prod_{b=1}^{n_B} P_{\mathcal{V}}(w_b/J) \\ \prod_{a=1}^{n_A} \int_{\mathcal{R}} P_{\mathcal{R}}(r_a/g_a)P_{G}(g_a/J)dg_a
\label{eq:crm}
\end{split}
\end{equation}
where $\mu$ is an empirically selected constant, $p_v$ is the relative frequency of $v$ in the training set, and $N_{v,J}$ is the number of times $v$ occurs in the observation $w_A$, and is used to get the Beysian estimate of the multinomial $P_{\mathcal{V}}(./J)$ \cite{CRM}.
\begin{equation}
P_{\mathcal{V}}(v/J) = \frac{\mu p_v + N_{v,J}}{ \mu + \sum_{v'}N_{v',J}}
\label{eq:vocab}
\end{equation}
Gaussian kernel is used for smoothing, while estimating $P_{G}(g/J)$ \cite{CRM}.
\begin{equation}
P_{G}(g_a/J) = \frac{1}{n}\sum_{i=1}^{n}\frac{\exp\{(g_a-G(r_i))^T|\Sigma|^{-1}(g_a-G(r_i))\}}{\sqrt{2^k\pi^k|\Sigma|}}
\label{eq:gauss}
\end{equation}
$P_\mathcal{T}(J)$ can be assumed constant and Lavrenko et al.~ have used a particularly simple expression for the distribution $P_{\mathcal{R}}(r/g)$.
\begin{equation}
P_{\mathcal{R}}(r/g) = \begin{cases} c , & \mbox{if } G(r) = g\\ 0 , & \mbox{otherwise } \end{cases}
\end{equation}
where $c$ is a constant independent of $g$. \\

Another relevance model was introduced by Feng et al.~ with the name of Multiple Bernoulli Relevance Model (MBRM). The difference between MBRM and CRM is that MBRM assumes a Bernoulli distribution of the vocabulary for $P_{\mathcal{V}}(v/J)$. Since it is rare for words to be repeated in the description of images, the Bernoulli distribution seems to be better suited to estimate $P_{\mathcal{V}}(v/J) $ with the following expression (\ref{eq:vocab2}), and the MBRM shows improvement over the CRM \cite{MBRM}.
\begin{equation}
P_{\mathcal{V}}(v/J) = \frac{\mu \delta_{v,J} + N_{v}}{ \mu +N}
\label{eq:vocab2}
\end{equation}

In equation (\ref{eq:vocab2}), $ \delta_{v,J}$ represents the presence/absence of a word in the annotations of image $J$, $N_v$ is the number of images in the training set containing $v$ as annotation, and $N$ is the total number of images in the training set.\\

We propose two modifications to adapt the CRM to image theme modeling based on the following two observations: (i) images of the same themes often have similar spatial coherence, (ii) images of the same themes often exhibit similar visual characteristics. The following two sections describe the modification of the CRM according to these two observations. To emphasize the spatial and visual coherence among images of common themes, we call this model a {\em Coherent Continuous Relevance Model} (CCRM).

\subsubsection{Relevance Model with Spatial Coherence}
\label{sn:CCRM}
Over the years, different methods have been explored to divide images in regions to generate $ \{r_1,r_2...r_{n_A}\}$ e.g. Blobworld \cite{bbworld} which is a complex method to divide an image in regions of similar appearance, which can generate different number of blobs in different images. In our proposed model, we want to generate equal number of regions in all images to preserve spatial coherence. Therefore, we have generated visual features by dividing images in a fixed grid and then representing each tile of grid with $30$ features representing color and texture of these tiles. Our features include $18$ color features: mean and standard deviation of RGB, LUV and LAB color components, and $12$ texture features: output of the Gabor filter with $3$ different scales and $4$ orientations. It has been previously proven that grid-based visual features perform better than complex blob based features \cite{MBRM}. \\

Using the same notation as used by Lavrenko et al., $r_a$ represents a region of image A and the function $G(r_a)$ produces $g_a$ as the corresponding visual feature (generator vector) for the region $r_a$.  In CCRM, the joint probability of observing $w_B$ with image $A$ is estimated by the same process of expectation over all images in the training set as described for CRM in equation (\ref{eq:crm}). \\

As argued earlier, the visual themes are not only captured by the visual features of tiles, but also by their relative spatial coherence. Therefore, we modify the equation for $P_{G}(g_a/J)$ (i.e. equation (\ref{eq:gauss})) to incorporate the spatial information of the visual features. If $a$ represents one tile at a certain position in all images (as all images have the same number of tiles),  $g_a$ represents the visual feature corresponding to the $a${\em  th} tile in image $A$, and $r_a$ represents the corresponding image region in image $J$ from the training set.

\begin{align}
P_{G}(g_a/J) = \frac{\exp\{(g_a-G(r_a))^T|\Sigma|^{-1}(g_a-G(r_a))\}}{\sqrt{2^k\pi^k|\Sigma|}}
\label{eq:gauss2}
\end{align}

In equation (\ref{eq:gauss2}), $P_G(g_a/J)$ depends only on the corresponding tile of image $J$. An additional advantage of this modification to CRM is the huge reduction in time-complexity of the procedure. The complexity is actually reduced by a factor of $n$, where $n$ is the number of tiles in the grid. To execute this approach, it is necessary that all images are divided by the same size grid, and that ordering of tiles are fixed in the representation of visual features.  Another point to be noted is that in the case of theme-based image annotation, each word $w_b$ is actually an image theme, which may be represented as $t_b$. Therefore, in estimating $P_{\mathcal{V}}(t_b/J)$ using the Dirichlet prior, the count of word $v$ is replaced by the strength of the image theme $t_b$.

\subsubsection{Clustering Based on Visual Features}
\label{sn:clustering}
Theme-based image annotation assumes the presence of significant visual themes in images. To take advantage of such common visual themes among multiple images in a data set, we incorporate a step of clustering images based on their visual features before image theme annotation. The underlying assumption is that there is visual coherence among images belonging to one cluster. We fixed the size of clusters and dropped clusters with too-low membership based on the assumption that those clusters represent images with rarely-occurring themes. Image theme modeling is applied over each cluster to generate image themes specific to that cluster only. For annotation, an image is matched to a suitable cluster based on its visual features, and then the annotation process is carried out using the information from that particular cluster only. Once an image is matched to its corresponding cluster, the pool of possible image themes for annotation is smaller than the total number of image themes for the complete data set. Even within that smaller pool of image themes, image themes share an underlying common context, since these image themes are shared by images with some level of visual coherence. \\

Equations provided in the previous section to estimate the probabilities are therefore modified based on this additional clustering step. Let the test image $A$ be matched to the cluster $C_m$, and let $\mathcal{T}_{C_m}$ be the set of all training images belonging to this cluster. The following equation represents our modified probability estimation:
\begin{equation}
\begin{split}
P(r_A,t_B) = \sum_{J \epsilon \mathcal{T}_{C_m}} P_{\mathcal{T}}(J) \prod_{b=1}^{n_B} P_{\mathcal{V}}(t_b/J) \\ \prod_{a=1}^{n_A} \int_{\mathcal{R}} P_{\mathcal{R}}(r_a/g_a)P_{G}(g_a/J)dg_a
\end{split}
\end{equation}

As mentioned earlier, $t_b$ represents the image themes instead of the words for theme-based image annotation.  $P_{\mathcal{V}}(t_b/J)$ is estimated over the samples of one particular cluster.

\section{Data Set and Results}
\label{sn:res}
Our main data set for evaluation is the IAPR TC 12 consisting of 20,000 images. Each image has accompanying description provided in three languages. At present, we have restricted our experiments to use only the English descriptions. We have conducted experiments to confirm the quality of each of the modification we have made to the relevance models. In this section, we shall present the details of these experiments.

\subsection{Evaluating Spatial Coherence}
The first set of experiments are used to demonstrate the validity of spatial coherence even within classical relevance models (i.e. MBRM \cite{MBRM} in this case). Makadia et al. have used the classical MBRM to annotate two rather challenging data sets i.e. the IAPR TC 12 and the ESP\footnote{http://www.espgame.org} game.  We ran MBRM, while preserving spatial coherence, and compared our results against those provided by Makadia et al. To generate comparable results, we used similar data set and vocabulary sizes. The number of annotations per image is the same as the average length of annotation per image in the training set. Our results in Table \ref{tb:esp} confirm that incorporating spatial information produces better results with much lower computational time, confirming thus the validity of the idea of spatial coherence. An additional advantage is the considerable reduction in computational complexity of the new method. Comparing equations (\ref{eq:gauss}) and (\ref{eq:gauss2}), it is evident that preserving spatial coherence in the relevance model reduces the complexity by an order of $n$, where $n$ is the total number of regions in each image.

\subsection{Evaluating CCRM Without Clustering}
For the remaining experiments to evaluate the CCRM, we used the IAPR TC 12 data set restricted to English descriptions. For this purpose, we generated image themes from the descriptions by using the Latent Dirichlet Allocation \footnote{http://psiexp.ss.uci.edu/research/programs\_data/toolbox.htm} with fixed number of image themes. As described in section \ref{sn:TM}, LDA takes as input a collection of documents, which in our case is the set of all image descriptions. LDA uses variational  inference algorithms to estimate word distributions conditioned over image theme distribution. It returns as output word distributions for a certain number of image themes. The number of image themes is a user input and therefore can be changed. Examples of these word distributions have been provided in table \ref{tb:image themes}. It also returns vectors indicating which image themes are present in each document, i.e. each image description. \\

For experimental purposes, we used LDA over the description of all images, and generated a vector indicating the presence or the absence of an image theme corresponding to each image. We, thus, converted our problem from the word domain to the theme domain, and generated the ground truth for evaluation. We then separated the test and the training sets and used CCRM to annotate the images in the test set with image themes, using the expectation over all images in the training data, as described in section \ref{sn:RM}. Note that annotations generated in this case are actually image themes and not words. Theme-based image annotations are useful, as we have already established that image themes corresponds to visual themes. Therefore, a correct theme-based image annotation means also a correct identification of visual themes. Later on, suchthis visual themes can be represented in terms of top few words from the word distribution corresponding to the image themes. These words may also be used to generate phrases if used with some natural language generation process. \\

We compared image theme annotation for test images generated through CCRM with the ground-truth generated by using LDA in the first step. We measured the performance of the proposed theme-based annotation in terms of mean precision, recall and F-measure per image theme. Results are provided in Table \ref{tb:ovt}, with different number of image themes generated through LDA.

\subsection{Evaluating CCRM With Clustering}
We conducted another set of experiments to apply the second modification to the classical relevance model that we suggested in section \ref{sn:clustering}. For these experiments, we first clustered the available images based on the similarity in their visual features. We used standard Euclidean distance as the measure of similarity between images for clustering and then dropped clusters with extremely low membership. We were left with about $80\%$ of the original data set and $46$ clusters. \\

For theme-based image annotation, we applied LDA over image descriptions in each cluster separately, and the number of image themes was decided based on the membership of that cluster. Typically, a cluster with larger number of images is likely to have images representing greater number of themes, therefore larger number of image themes. Again, the test and the training sets were separated after generating ground truth in the theme-domain. CCRM was then applied to estimate the joint probability of visual features and themes for each of the test images, using only the training images from its corresponding cluster. \\ 

Results are provided in Table \ref{tb:ct}. Image themes generated for each cluster are treated as distinct from image themes generated for other clusters. Therefore, a large number of image themes are generated, which makes the theme space more fine-grained. Results show improvement over experiments described in the previous section even though the theme-space is now larger and more fine-grained. the performance depends also on the number of image themes generated. Overall, clustering helps the process of identification of correct visual themes and the association of images with the corresponding themes. \\

As described earlier, image themes are represented by their word-distribution. No two image themes generated will have the exact same word-distribution. If a large number of image themes are generated, then the average similarity between image themes increases, i.e. some image themes share similar but not equivalent word distributions.

\subsection{Image Theme Annotation}
We annotated images with the image theme number (label), because image themes correspond to visual themes and we assumed that each theme is sufficiently described by a few top words or word distribution conditioned over that image theme, generated using LDA. The top few words of the word distribution of each image theme are conceptually highly related. All the words assigned to an image, based on the word distributions of themes for that image, may not be present in the original description of the image. However, the additional words are commonsense extensions to already provided descriptions by users. For example, ``tree'', ``trunk'',  and ``leaves'' are top three words of the word distribution of an image theme, and they have been assigned to an image that has a `tree' in it, but the original image description did not contain the word ``trunk''. Standard precision and recall measures will consider ``trunk'' as wrong annotation, while it is actually a conceptually related detail to the visual theme of ``tree''.

\subsection{Comparison With Other Methods}
We ran another set of experiments to compare our results against state-of-the-art algorithms for keyword annotation. Although, image theme annotation is a higher-level semantic annotation, we can compare the results by treating the top words form the word distribution of image themes as keywords. Table \ref{tb:wds} shows that, on comparable problem sizes, our theme-based image annotation beats the CRM with spatial coherence, the MBRM, and the two greedy label transfer algorithms described by Makadia et al.~ \cite{baseline}, i.e. JEC and Lasso. To make the results comparable, we used the same sizes of data set.  Makadia et al.~ have used $291$ most frequently occurring keywords as vocabulary. We ran CRM with spatial coherence using the same vocabulary size. We annotated the images with image themes using CCRM with clustering and used the top $3$ words of image themes assigned to the image with highest probability as the annotating keywords. There were a total of $360$ distinct words appearing in annotation in this case. Results in Table \ref{tb:wds} are expressed in terms of precision and recall, averaged over the total number of keywords used in the experiment, making thus the numerical results comparable. \\

The improvement generated by our method is even more significant considering the fact that the state-of-the-art relevance models were beaten in performance by the simple greedy approach algorithms \cite{baseline}. Many of the variations suggested over time use computationally expensive approaches for restricted data sets like Corel5K, where image contents basically belong to a few broad categories, e.g. `cars' , `animals', etc. Methods using language models or ontology like WordNet enjoye limited success. We have used a larger dataset with significant variation in image contents. Our method shows not only performance improvement over other relevance models, but also over the greedy label transfer based algorithms.

\begin{table*}[htbp]
\centering
\caption{MBRM vs. MBRM with spatial coherence; Results of typical MBRM (described by Feng et a.l \cite{MBRM}) have been taken from Makadia et al. \cite{baseline}, which does not report F-measure values. Similar vocabulary and dataset size have been used in both cases, as used by Makadia et al. }
\label{tb:esp}
\begin{tabular}{|c|c|c|c|c|c|c|}\hline
\multirow{2}{*}{}Dataset & Annotation algorithms & Mean precision & Mean recall & Mean F-measure & No. of keywords & Total no. of \\ & &per keyword&per keyword&per keyword& with recall $>$ 0 & keywords\\\hline
\multirow{2}{*}{}ESP & MBRM &21.0\%&17.0\%&--&218&268\\\hline
ESP & MBRM-spatial coherence&25.0\%&17.0\%&19.0\%&235&268\\\hline
IAPR &  MBRM&21.0\%&14.0\%&--&186&291\\\hline
IAPR &  MBRM-spatial coherence &24.0\%&15.0\%&16.0\%&213&291\\\hline
\end{tabular}
\end{table*}

\begin{table*}[htbp]
\centering
\caption{Image theme annotation performance }
\label{tb:ovt}
\begin{tabular}{|c|c|c|c|c|}\hline
\multirow{2}{*}{} & Mean precision & Mean recall & Mean F-measure & Number of themes \\ &per theme&per theme&per theme& with recall $>$ 0\\\hline
25 themes over all dataset&44.2\%&35.0\%&33.0\%&25\\\hline
50 themes over all dataset&31.0\%&24.0\%&22.6\%&50\\\hline
75 themes over all dataset&26.3\%&19.0\%&18.0\%&75\\\hline
\end{tabular}
\end{table*}

\begin{table*}[htbp]
\centering
\caption{Image theme annotation performance with clustering; number of image themes for each cluster is decided on the basis of number of members of cluster}
\label{tb:ct}
\begin{tabular}{|c|c|c|c|c|c|}\hline
\multirow{2}{*}{} & Mean precision & Mean recall & Mean F-measure & Number of themes & Total number of \\ &per theme&per theme&per theme& with recall $>$ 0 & themes\\\hline
2 themes per 100 images&62.2\%&57.1\%&55.0\%&323&327\\\hline
3 themes per 100 images&51.0\%&47.0\%&44.0\%&472&498\\\hline
4 themes per 100 images&44.0\%&40.3\%&37.2\%&601&671\\\hline
\end{tabular}
\end{table*}

\begin{table*}[htbp]
\centering
\caption{Keyword annotation performance comparison between CRM-spatial coherence , MBRM \cite{baseline}, Lasso \cite{baseline} , JEC \cite{baseline} and words generated from image annotation with image themes using CCRM with clustering. All results have been computed over $20000$ images of IAPR dataset; with almost $10\%$ of dataset used as testing set. Results have been averaged out over total number of keywords as explained in section \ref{sn:res}. Results of MBRM, JEC and Lasso have been taken from Makadia et al., who have not reported F-measure \cite{baseline}. }
\label{tb:wds}
\begin{tabular}{|c|c|c|c|c|c|c|}\hline
\multirow{2}{*}{} & Mean precision & Mean recall & Mean F-measure & No. of words & Total no. of \\ &per word&per word&per word& with recall $>$ 0&words\\\hline
MBRM&21.0\%&14.0\%&--&186&291\\\hline
JEC&25.0\%&16.0\%&--&196&291\\\hline
Lasso&26.0\%&16.0\%&--&199&291\\\hline
CRM-spatial coherence&26.0\%&13.0\%&14.0\%&181&291\\\hline
Words generated from image themes &35.0\%&16.0\%&20.0\%&253&360\\\hline
\end{tabular}
\end{table*}

\begin{table*}[htbp]
\centering
\caption{Average relatedness score of top few words of $50$ image theme generated through image theme modeling over complete IAPR TC 12 dataset}
\label{tb:cnconf}
\begin{tabular}{|c|c|c|c|}\hline
No. of top words & Avg. score for all themes & Avg. score for best 10 themes & Avg. score for best 5 themes\\\hline
5&28.1&37.5&41.3\\\hline
6&27.8&36.3&38.5\\\hline
\end{tabular}
\end{table*}

\subsection{Ground Truth Augmentation Using ConceptNet}
Our assumption is that images annotated with image themes provide useful and understandable information to humans. To prove our point, we employed ConceptNet. We explained in details in section \ref{sn:CN} that connections between nodes are weighted in the semantic network of ConceptNet. The relatedness score provided by ConceptNet takes into account all paths between two nodes in the network and also their weights. Heavier weights mean more conceptual relatedness between nodes. Table \ref{tb:cnconf} provides a few statistics of the relatedness scores between the top few words of each image theme. We estimated that average numerical relatedness score of top $5$ words of all image themes was about $28\%$ \footnote{This estimation was performed over 50 image themes generated over the entire data set}. \\

We explained in section \ref{sn:CN} the difference between image themes generated through LDA and concepts present in ConceptNet. Our observations indicate that the top few words of each image theme convey a strong, commonsense acceptable hint about image content to humans. There can be two ways to represent an image annotated with an image theme. One simple way is to annotate it with a fixed number of top words from word distributions of the selected image themes. Although, these words may not describe all objects in the image, they would definitely provide a strong indication of the image context. The second way is to generate a phrase encompassing the top words of each image theme to annotate the image. For example, top few words of one of the image themes were ``sand'', ``beach'', and ``water'', for which ``sandy beach'' could be an appropriate annotation. Phrase generation is a problem that has been studied in NLP. \\

Additional advantage of using image theme modeling with the aid of ConceptNet is that we can augment our ground-truth for a given annotated data set. ConceptNet can be used to elaborate at least those image themes that are present in the image. ConceptNet provides the possibility of extending a word list provided as input with conceptually related words, which is called `projection' within the context of ConceptNet. The API of ConceptNEt 2.0 has this facility with the same name. This API even provides distinction between different types of projections e.g. spatial projections, where projected words have spatial relations to input words, e.g. California is part of spatial projection of Los Angeles \cite{CN}. Other types of projections include consequence, detail, etc., where projected words are consequence or part of detailed descriptions of input words, respectively. The output list is a list of words, sorted according to their aggregate relatedness scores with the words in input list. We have used this API to augment the annotation generated by CCRM and image theme modeling. We first annotated each image with top few words of the image themes found using CCRM. Then we provided the list of annotation as input to the `projection' facility of the ConceptNet API, and found an extended list of conceptually related words. We argue that using top words from these extended list as annotations for images provides an even more detailed description of images. The words provided by `projection' may not have been used by user as the descriptions of the images in our test data set. However, these words are clearly conceptually related to the visual themes of the images. Therefore, these words were used to augment the ground truth information i.e. image descriptions provided by users. We have included some examples to demonstrate this fact. \\

\begin{figure*}%[htbp]
\centering
\includegraphics[height=2.5in , width=7.0in]{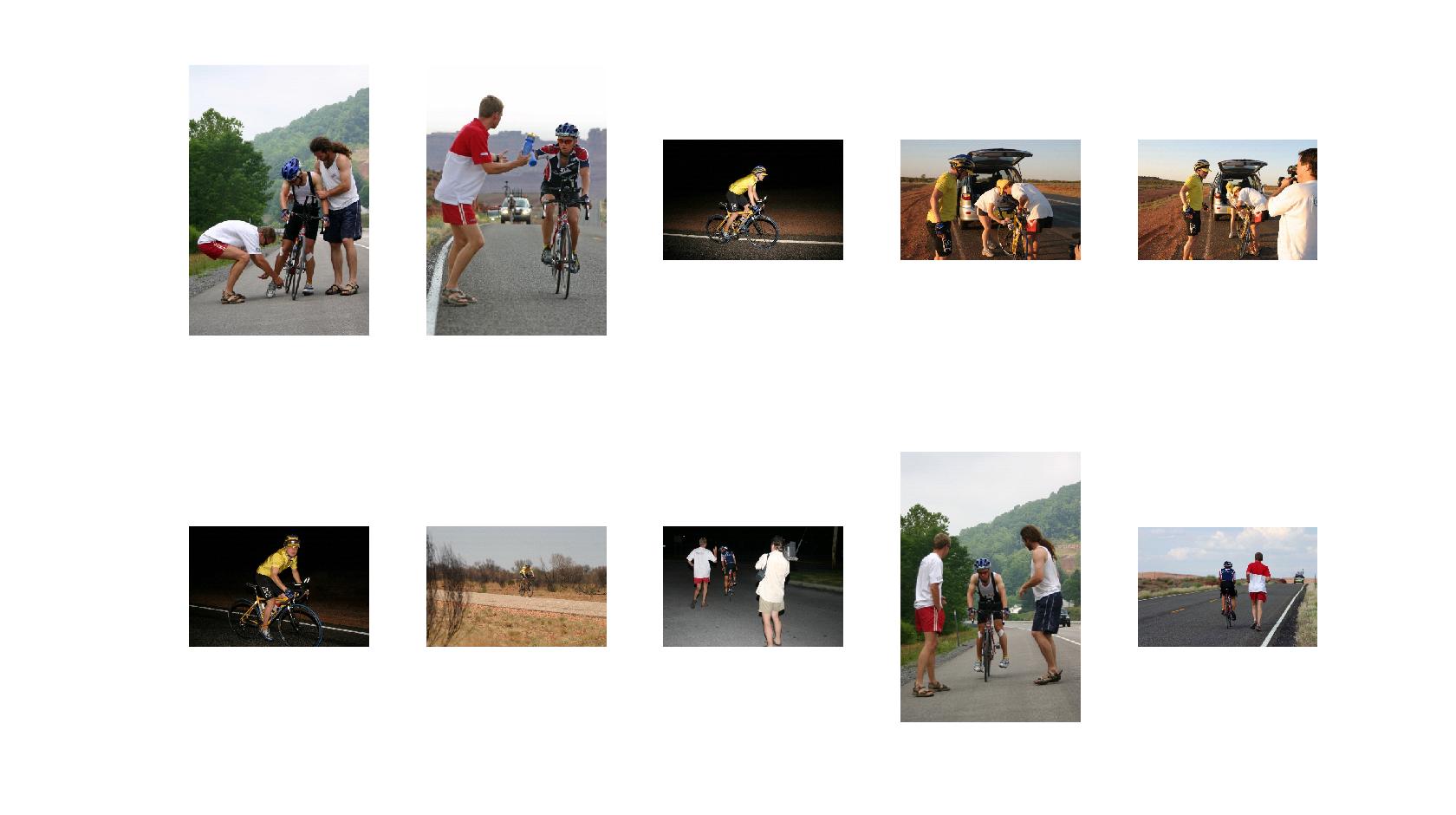}
\caption{Top words from image theme : `shorts'  `cyclist' , `helmet' , `jersey' ; Augmented words: `bike-ride' , `athlete'}
\label{fg:bike_ride}
\end{figure*}

\begin{figure*}%[htbp]
\centering
\includegraphics[height=2.5in , width=7.0in]{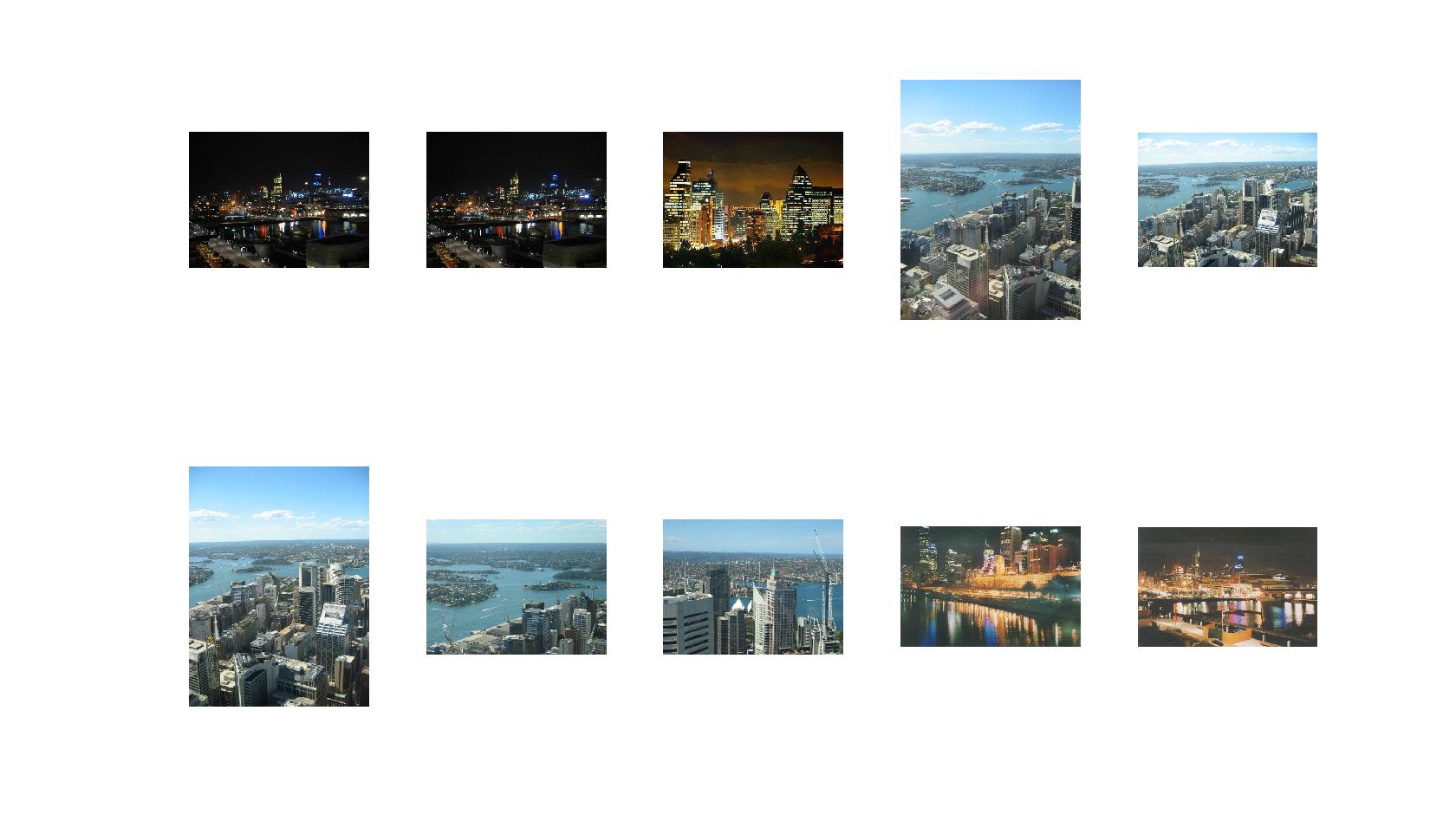}
\caption{Top words from image theme :`building' , `city' , `view' , `night' ; Augmented words: `look-through-telescope' , `dark'}
\label{fg:telescope}
\end{figure*}

\begin{figure*}%[htbp]
\centering
\includegraphics[height=2.5in , width=7.0in]{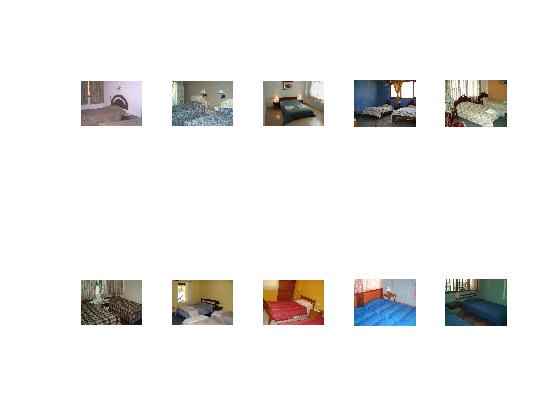}
\caption{Top words from image theme :`room' , `wood' , `walls' , `lamp' ; Augmented words: `furniture'}
\label{fg:furniture}
\end{figure*}

\begin{figure*}[htbp]
\centering
\includegraphics[height=2.5in , width=7.0in]{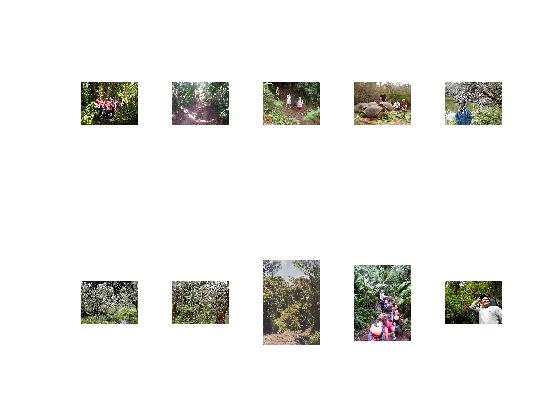}
\caption{Top words from image theme :`forest' , `bushes' , `dense' , `path' ; Augmented words: `nature' }
\label{fg:nature}
\end{figure*}

Table \ref{tb:cn} provides an example list of top words from the word distributions conditioned over a few image themes in our data set, and useful projections provided by ConceptNet. Another observation that we made was that ConceptNet can provide more abstract ideas in projected words, e.g. `nature' is a word projected for images of forest. When added to the description of an image, this word can appropriately make the image relevant to queries dealing with the general idea of `nature'.
\begin{table*}[htbp]
\centering
\caption{Sample of top words of a few image themes and augmented words from ConceptNet; These image themes have been selected from $50$ image themes generated over all dataset }
\label{tb:cn}
\begin{tabular}{|c|c|}\hline
\textbf{Top words of image theme} & \textbf{Augmented words using ConceptNet}\\\hline
`walls' , `children' , `classroom' , `board' , `desk'&`in-school' , `student'\\\hline
`clouds' ,`sky' , `sun' , `shade'&`sunset' , `yellow' , `blue'\\\hline
`forest' , `bushes' , `dense' , `path'&`nature' \\\hline
`spectators' , `stadium' , `grandstand' , `court'&`game-play' , `watch' \\\hline
`room' , `wood' , `walls' , `lamp'&`furniture'\\\hline
`building' , `city' , `view' , `night'&`look-through-telescope' , `dark'\\\hline
`road' , `gravel' , `car' , `dirt'&`ride' , `track'\\\hline
`streets' , `building' , `people' , `pavement'&`road' , `walk' ,`in-city'\\\hline
`shorts'  `cyclist' , `helmet' , `jersey'&`bike-ride' , `athlete'\\\hline
`man' , `woman' , `shirt' , `hands' , `clothes'&`wear' , `outfit' , `dress'\\\hline
\end{tabular}
\end{table*}
Figures \ref{fg:bike_ride} \ref{fg:telescope} \ref{fg:furniture} \ref{fg:nature} show sample of images annotated with a certain image theme; with top few words from word distribution of the image theme and additional projected words provided in caption. It is evident that projected words , if augmented to ground-truth, provide useful context about images. Space restrictions prohibit us from providing more samples.

\begin{figure*}[htbpp]
\centering
\includegraphics[height=2.5in , width=7.0in]{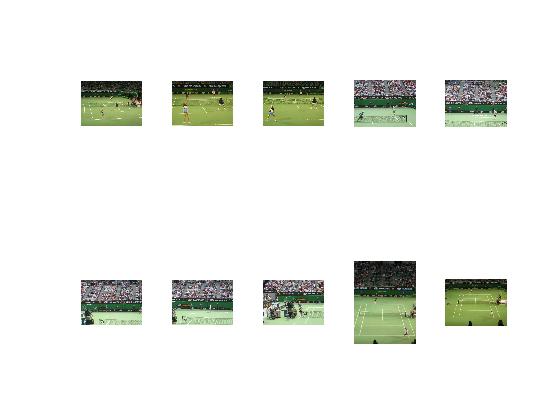}
\caption{Top words from image theme :`spectators' , `stadium' , `grandstand' , `court'; Augmented words: `game-play' , `watch' }
\label{fg:game}
\end{figure*}

\newpage

\section{Conclusion}
Automatic image annotation has been a focus of research because of its potential application to benefit image search and retrieval engines, as well as many other applications in image/video processing. Most of the algorithms presented previously perform unsatisfactorily when tested over challenging data sets like IAPR TC 12. We have radically transformed the problem of automatic image annotation from the keyword space to the image theme space. We have employed techniques popular in natural language processing (NLP) to annotate images with image themes corresponding to visual themes rather than independent keywords corresponding to individual objects. Annotated images, when represented by a few significant words from the word distribution of the image themes, can provide strong conceptually-acceptable hint towards the overall theme of the image.  We have employed for the first time a semantic network (i.e. the ConceptNet) to provide commonsense basis of our image theme annotation idea. We have also shown that matching an image to a cluster of images with similar visual themes helps narrow down possible image themes for annotation, while providing a performance boost. By using top words from the word distribution of the image themes as annotations, we have compared the performance against standard keyword-based annotation methods, and have shown superior results. Performance improvement is even more significant considering the fact that most previously developed methods have been beaten by the greedy label transfer approaches, when used for a challenging data set like IAPR - our system is able to beat these greedy approaches. 

\clearpage
\bibliographystyle{plain}
\bibliography{foroosh,references}

\end{document}